# DELTA-POSITION ESTIMATION-BASED IMU ODOMETRY: A COMPARISON OF MLP AND KOLMOGOROV–ARNOLD NETWORKS

Osman Tokluoğlu[1]*, Emin Keresteci[2]

[1]Department of Electrical and Electronics Engineering, Ankara Yıldırım Beyazıt University, Ankara, Türkiye

[2]Department of Computer Engineering, TOBB University of Economics and Technology, Ankara, Türkiye

**Abstract:** In this study, the learning-based inertial odometry problem is investigated using raw IMU measurements obtained from the EuRoC MAV benchmark dataset. Instead of absolute position regression—a formulation that may lead to large constant errors—the models are trained to estimate the incremental displacement (Δp) over a fixed 50 ms sliding window, and the full trajectory is reconstructed through numerical integration. A standard Multi-Layer Perceptron (MLP) is compared with a Kolmogorov–Arnold Network (KAN) equipped with learnable B-spline activations. Although KAN has 6.9 times fewer parameters than MLP (8,444 versus 57,859), it produces a 44% lower error in terms of final cumulative drift on the test trajectory (9.61 m versus 17.23 m). In addition, KAN exhibits more stable behavior in terms of long-term error accumulation, with lower $P_{50}$ and $P_{90}$ cumulative drift values. These findings indicate that learnable B-spline-based activations have the potential to reduce error accumulation in the inertial odometry problem.

## INTRODUCTION

Accurate and reliable position estimation for autonomous aerial vehicles in GPS-denied indoor environments remains a fundamental challenge. Inertial Measurement Units (IMUs) offer an attractive primary sensing modality by providing high-frequency measurements of angular velocity and specific force. However, IMU-based dead reckoning suffers from the well-known problem of error accumulation: small biases and noise in raw measurements are integrated twice to obtain position, leading to drift that grows quadratically over time under classical mechanization equations (Woodman, 2007).

Classical Inertial Navigation System (INS) approaches attempt to address this issue through tight fusion with complementary sensors such as cameras, LiDARs, or barometers, together with model-based filtering techniques such as the Extended Kalman Filter or more recent factor graph optimization methods. Although these methods can provide centimeter-level accuracy under controlled conditions, their performance rapidly degrades when sensor models become mismatched, IMU biases vary over time, or the underlying kinematic assumptions of the filter no longer hold (Forster et al., 2017). Moreover, model-based approaches are often platform- and environment-specific, requiring careful tuning of noise covariance parameters.

Learning-based approaches have emerged as a promising alternative due to their ability to capture complex and nonlinear sensor error characteristics directly from data. Pioneering studies such as TLIO (Liu et al., 2020) have shown that a convolutional network trained on IMU windows can estimate displacement increments with competitive accuracy. IONet (Chen et al., 2018) and RONIN (Herath et al., 2020) extended this paradigm to pedestrian motion using recurrent architectures. A common design choice in these methods is the formulation of the problem as local displacement regression, which avoids the drift associated with absolute position regression while keeping trajectory reconstruction through numerical integration practical.

Meanwhile, Kolmogorov–Arnold Networks (KANs) have been proposed as a theoretically motivated alternative to standard MLPs (Liu et al., 2024). Based on the Kolmogorov–Arnold representation theorem, KANs place learnable univariate B-spline functions on the edges of the network rather than using fixed activation functions at the nodes, enabling compact yet highly expressive models. Initial results suggest that KANs can match or even surpass the accuracy of MLPs in scientific regression tasks with substantially fewer parameters. However, their applicability to high-frequency time-series regression problems—such as IMU odometry—has not yet been systematically evaluated.

In this paper, we present a controlled and empirical study comparing MLP and KAN architectures for inertial odometry on the EuRoC MAV benchmark dataset (Burri et al., 2016). Our contributions are threefold. First, we show that a KAN with only 8,444 parameters achieves 44% lower final cumulative drift than a 57,859-parameter MLP on the same sliding-window delta-position estimation task. Second, we present systematic ablation studies over the IMU sampling frequency (25–200 Hz) and window size (W = 5–50), identifying the operating conditions under which each architecture performs best. Third, we conduct zero-shot cross-sequence evaluation and provide an interpretability analysis of the learned KAN spline functions, revealing physically meaningful channel preferences and locality biases that are consistent with inertial sensor physics.

## LITERATURE REVIEW

Traditional INS mechanization integrates specific force and angular velocity measurements in a selected reference frame using the Newton–Euler equations (Titterton & Weston, 2004). Without auxiliary sensors, position error grows proportionally to the cube of time due to the double integration of accelerometer noise. Practical systems address this issue by fusing IMU data with GPS, magnetometers, or visual information through Kalman-type filters. The Multi-State Constraint Kalman Filter (MSCKF) of Mourikis and Roumeliotis (2007) and OKVIS (Leutenegger et al., 2015) are prominent examples of model-based approaches in visual–inertial odometry; however, they remain dependent on camera availability and well-calibrated sensor models specific to each platform.

TLIO (Liu et al., 2020) introduced a ResNet architecture trained to estimate 3D displacement and its uncertainty from 1-second IMU windows and achieved sub-meter accuracy over 60-second corridor trajectories when combined with an EKF. RONIN (Herath et al., 2020) compared TCN, LSTM, and Transformer backbones on pedestrian IMU data and found that temporal convolutional networks provided the best balance between accuracy and latency. AbolDeepIO (Esfahani et al., 2019) applied deep networks to UAV inertial odometry and reported drift reductions of up to 30% compared with preintegrated IMU in a VIO pipeline. The delta-position regression formulation, which is a common design choice in these methods, avoids the drift associated with absolute position regression.

Liu et al. (2024) proposed KANs, inspired by the Kolmogorov–Arnold representation theorem, which states that any multivariate continuous function can be represented as a composition and summation of univariate functions. Specifically, KANs parameterize each edge function $\varphi_{ij}$ as a learnable B-spline plus a residual SiLU basis and place these functions on the edges rather than at the nodes. The authors reported that, in physics-informed regression tasks, KANs achieved accuracy comparable to MLPs with 10–100× fewer parameters while producing interpretable symbolic function graphs. Subsequent studies have investigated KANs for time-series classification (Xu et al., 2024) and medical image segmentation (Li et al., 2024); however, to the best of our knowledge, no prior study has evaluated KANs for high-frequency IMU odometry. Although Keresteci et al. (2025) evaluated KANs in a spatio-temporal mobility-based user identification setting, this study aims to provide a more comprehensive assessment of the potential of the KAN architecture for IMU odometry.

## DATASET

This section describes in detail the EuRoC MAV dataset used in the experiments and the data preparation process. First, the selected flight sequence, IMU sensor measurements, and Vicon-based ground-truth reference information are introduced. Then, the steps of timestamp-based alignment, chronological data splitting, normalization, and sliding-window generation are discussed. Finally, the incremental position-change formulation targeted by the model and the final dataset dimensions are presented.

### EuRoC MAV Benchmark Dataset

All experiments use the EuRoC Micro Aerial Vehicle (MAV) dataset (Burri et al., 2016). In this study, the V1_01_easy sequence recorded in the Vicon Room 1 indoor environment is specifically selected. Since this sequence provides a continuous and well-calibrated flight at moderate speed in a 6 m × 6 m room, it constitutes a clean reference case for analyzing drift accumulation over time.

The platform is equipped with an ADIS16448 MEMS IMU sensor operating at a sampling frequency of 200 Hz. Each IMU sample consists of six components expressed in the sensor body coordinate frame: three-axis angular velocity components ($\omega_x$, $\omega_y$, $\omega_z$) in $rad \cdot s^{-1}$ and three-axis specific force/acceleration components ($a_x$, $a_y$, $a_z$) in $m \cdot s^{-2}$.

The position and orientation ground-truth values are provided by a Vicon motion capture system operating at approximately 200 Hz. As the regression target, only the position vector in meters, $p = (p_x, p_y, p_z)^T$, is used.

### Data Alignment and Preprocessing

The IMU and ground-truth data streams contain hardware timestamps in nanoseconds. Alignment is performed by first sorting both streams independently and then applying a nearest-neighbor merge based on the timestamp column. After the merge, rows containing missing values are discarded, resulting in 29,120 synchronized samples covering the complete flight.

### Temporal Train/Validation/Test Split

To prevent information leakage, all splits are performed before any normalization step. The 29,120 samples are chronologically divided as follows: training + validation = the first 80% (23,296 samples), and test = the final 20% (5,824 samples). A StandardScaler is fitted only to the IMU channels in the training split and then applied to the validation and test data.

### Sliding-Window Feature Generation

The raw IMU measurements are arranged into sliding windows of length W = 10 samples. Since the IMU data have a sampling frequency of 200 Hz, each window corresponds to an approximately 50 ms time interval. Each generated window, $x_t \in \mathbb{R}^{10x6}$, is converted into a 60-dimensional feature vector for use in MLP-based models, while it is preserved in its 10 × 6 matrix form for reference models that exploit the sequential data structure.

### Delta-Position Formulation

In this study, the regression target is defined as incremental displacement rather than absolute position. Accordingly, for each window, the target value is computed as the position difference between the start and end instants of the window: $\Delta p_t = p(t+W-1) - p(t)$. Here, $\Delta p_t$ represents the three-dimensional position change in meters over the 50 ms window. The standard deviations of the displacement components obtained from the training set are calculated as $\sigma_{\Delta x}$ = 0,012 m, $\sigma_{\Delta y}$ = 0,017 m, and $\sigma_{\Delta z}$ = 0,005 m, respectively.

### Final Dataset Dimensions

At the end of the data preparation process, the final number of windows to be used for model training and evaluation is obtained for each split. Due to the sliding-window structure with W = 10, the number of windows is lower than the number of raw samples. Table 1 presents the final dataset dimensions for the training, validation, and test sets.

Table 1. Number of windows after sliding-window generation.

| Split | Windows | Raw Samples |
|---|---|---|
| Training | 18,629 | 18,638 |
| Validation | 4,658 | 4,667 |
| Test | 5,815 | 5,824 |

## METHOD

### Problem Formulation

Given a normalized IMU window, $\tilde{x}_t \in \mathbb{R}^{60}$, the model $f_\theta$ is trained to minimize the mean squared error over the normalized delta-position targets. All metrics are computed after the predictions are transformed back into meters.

### Multi-Layer Perceptron (MLP)

The MLP baseline follows a standard deep feedforward architecture: $\mathbb{R}^{60} \rightarrow 256 \rightarrow 128 \rightarrow 64 \rightarrow \mathbb{R}^3$. Each hidden layer is followed by Batch Normalization, ReLU activation, and Dropout with p = 0.1. The total number of trainable parameters is calculated as 57,859.

### Kolmogorov–Arnold Network (KAN)

The Kolmogorov–Arnold Network (KAN) architecture (Liu et al., 2024) uses learnable univariate B-spline functions defined on each edge of the network, instead of the fixed activation functions used in conventional neural networks. This structure enables flexible input-dependent functions to be learned on each connection, rather than relying on linear weights and node-based activations. For an edge from node i to node j, the edge function corresponding to the input x_i is expressed as follows:

$$\phi_{ij}(x_i) = w_b b(x_i) + w_s \sum_k c_k B_k(x_i).$$

Here, $b(x) = x/(1 + e^{-x})$ denotes the residual SiLU basis function, while $B_k(x)$ represents the B-spline basis functions. The parameters w_b and w_s are learnable coefficients that control the contribution of the corresponding components, whereas $c_k$ denotes the learnable coefficients associated with the B-spline bases.

The KAN architecture used in this study has a width of [60, 7, 5, 3]. Cubic B-splines with k = 3 and a grid structure with 5 grid points are employed in the model. This structure contains a total of 8,444 trainable parameters, which is approximately 6.9 times fewer than the MLP model used for comparison.

### Training Procedure

Both models were trained under the same training protocol to ensure a fair comparison. The Adam algorithm was used for optimization. The learning rate was set to η_MLP = $10^{-3}$ for the MLP model and η_KAN = $5 \times 10^{-4}$ for the KAN model. The learning rate was updated using the ReduceLROnPlateau strategy, which monitors stagnation in the validation error. In this setting, the reduction factor was set to 0.5, the patience value to 20 epochs, and the minimum learning rate to $10^{-5}$.

To prevent overfitting and terminate the training process efficiently, an early stopping mechanism was employed. Training was stopped when no improvement greater than $10^{-5}$ was observed in the validation MSE for 80 consecutive epochs. The maximum number of epochs was set to 2000, and the random seed was fixed at 42 in all experiments. Under this training protocol, the MLP model converged at epoch 128 with MSE_val = 1.133, whereas the KAN model converged at epoch 332 with MSE_val = 1.256.

### Trajectory Reconstruction and Drift Metric

The predicted delta vectors, $\Delta p_t \in \mathbb{R}^3$, are integrated to reconstruct the full trajectory: $\hat{p}(0) = p^*(0)$, $\hat{p}(t) = \hat{p}(t-1) + \Delta p_t$ for $t \geq 1$. The cumulative drift is defined as $d(t) = \|\hat{p}(t) - p^*(t)\|_2$. The reported final drift corresponds to d(T) at the last test sample T.

### Evaluation Metrics

Four main metrics are used for performance evaluation. dRMSE denotes the root-mean-square value of the three-dimensional displacement error per step, expressed in meters. dMAE denotes the mean absolute value of the three-dimensional displacement error per step. To evaluate cumulative drift behavior, $P_{50}$ and $P_{90}$ metrics are used, where $P_{50}$ represents the median cumulative drift and $P_{90}$ represents the 90th percentile of cumulative drift. Finally, the final drift metric indicates the total accumulated position error at the last sample of the test sequence.

## EXPERIMENTS AND RESULTS

This section presents a comparative evaluation of the proposed KAN-based model and the reference MLP model on the test set. The evaluation is conducted using metrics that consider both per-step displacement prediction errors and cumulative drift behavior accumulated over time. Thus, the models are analyzed not only in terms of short-term prediction accuracy but also in terms of their ability to limit error accumulation in long-term position estimation. Table 2 summarizes the test results obtained on the V1_01_easy sequence using a window length of W = 10 and the delta-position formulation.

**Table 2.** Test-set performance on V1_01_easy (W = 10, delta formulation).

| Model | Params. | dRMSE (m) | dMAE (m) | $P_{50}$ (m) | $P_{90}$ (m) | Final drift (m) |
|---|---|---|---|---|---|---|
| MLP | 57,859 | 0.023 | 0.021 | 5.18 | 10.06 | 17.23 |
| KAN | 8,444 | **0.022** | **0.019** | **3.22** | **6.96** | **9.61** |

As shown in Table 2, the MLP and KAN models produce very similar results in terms of per-step displacement prediction errors. In particular, the difference between the dRMSE values is only 0.001 m, indicating that the two models exhibit comparable short-term prediction accuracy. However, when cumulative error behavior is examined, the KAN model provides a clear advantage. The final drift value of KAN is approximately 44% lower than that of the MLP model. This result indicates that the small step errors produced by the MLP accumulate more systematically over time, leading to larger position drift. In contrast, the learnable B-spline-based activations of KAN appear to produce error components that partially compensate for one another during the integration process.

In terms of computation time, the KAN model requires a more costly training process. While the MLP model is trained in approximately 41 seconds, the training time for the KAN model increases to approximately 460 seconds. Nevertheless, the fact that KAN has only 8,444 trainable parameters and offers an approximately 6.9 times more compact structure than MLP provides an important advantage in terms of model complexity. Therefore, despite its longer training time, KAN constitutes a notable alternative for compact motion-estimation applications requiring improved long-term stability, owing to its lower final drift and smaller parameter count.

To visually support the numerical results presented in Table 2, the position estimates and cumulative error behavior obtained throughout the test sequence are further examined. Figure 1 shows the predicted trajectories of the MLP and KAN models in the horizontal plane, together with the three-dimensional position errors accumulated over the time steps. This visualization enables the models to be compared not only in terms of total error values but also in terms of the temporal characteristics of error accumulation.

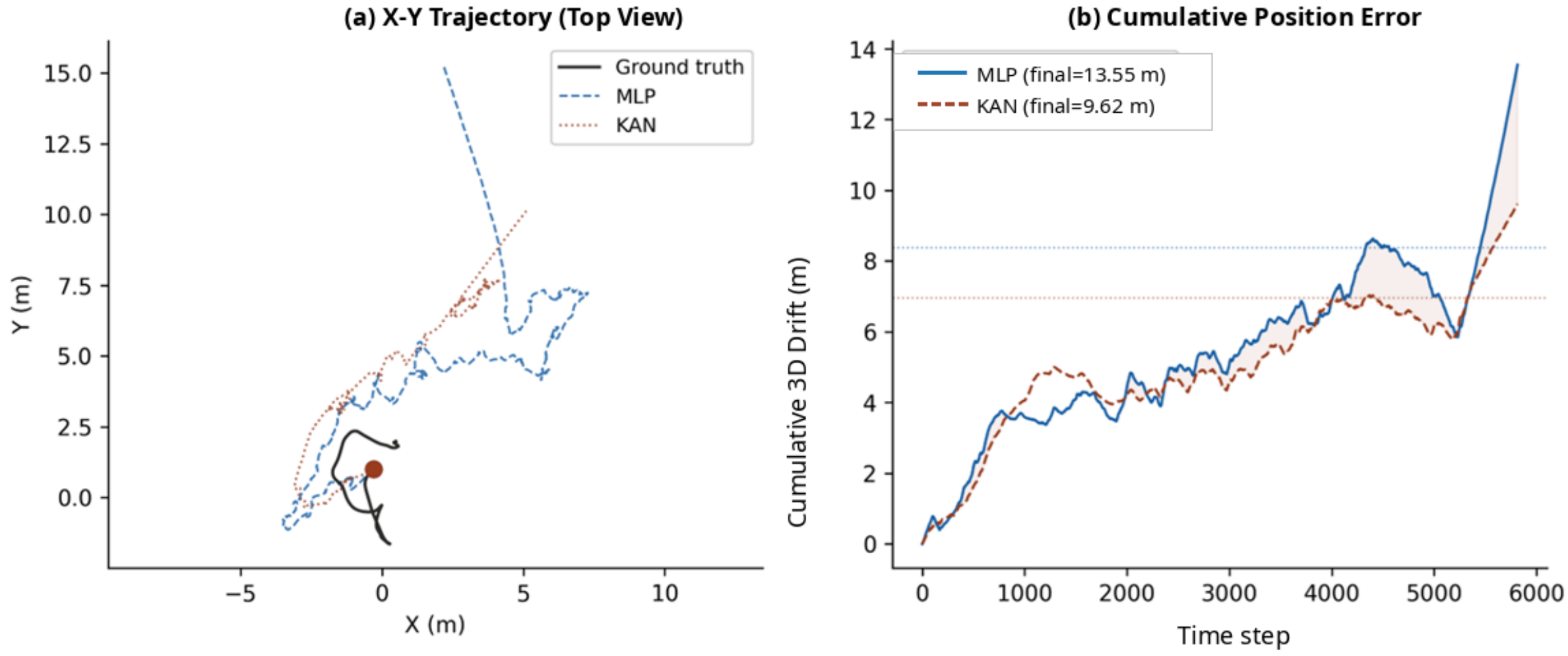

**Figure 1.** (a) Top-view X–Y trajectory. The ground-truth path is shown in black, MLP in blue, and KAN in dark red. MLP exhibits a pronounced monotonically increasing lateral deviation, whereas KAN remains closer to the ground truth. (b) Cumulative 3D drift. KAN has $P_{90}$ = 6.96 m, while MLP has $P_{90}$ = 10.06 m; the final drift is 9.61 m for KAN and 17.23 m for MLP. The shaded gap widens continuously after the first 5 s, indicating that the learnable activations of KAN provide a structural advantage rather than a merely transient effect.

## CONCLUSION

In this study, the performances of MLP and KAN architectures on the IMU-based inertial odometry problem were comparatively investigated using the EuRoC MAV benchmark dataset. The task was formulated as sliding-window delta-position regression rather than absolute position regression, and the predicted incremental displacements were combined through numerical integration to reconstruct the trajectory. This approach enabled the evaluation of not only the step-wise prediction accuracy of the models but also the cumulative position error accumulated over time.

The experimental results show that the KAN model can produce more successful results in terms of cumulative drift despite offering a more compact structure than MLP. In the configuration used, the KAN model has 8,444 trainable parameters, whereas the MLP model contains 57,859 parameters. Nevertheless, KAN reduces the final cumulative drift on the test trajectory by approximately 44% compared with MLP; the final drift is 17.23 m for MLP and 9.61 m for KAN. In addition, the KAN model produces lower values for the $P_{50}$ and $P_{90}$ cumulative drift metrics and exhibits more stable behavior in terms of long-term error accumulation.

When the per-step error metrics are examined, the MLP and KAN models are observed to yield very similar results. In particular, the difference between their dRMSE values is only 0.001 m, indicating that both models reach similar accuracy levels in short-term displacement estimation. However, the clear difference in cumulative error behavior reveals that how small step errors accumulate over time is critical for model performance. In this context, the learnable B-spline-based activation structure of KAN can be considered a notable alternative in terms of its potential to reduce error accumulation.

On the other hand, the KAN model is observed to be more costly than MLP in terms of computation time. While the MLP model is trained in approximately 41 seconds, the KAN model converges in approximately 460 seconds. Therefore, although KAN offers the advantages of a lower parameter count and lower final drift, it introduces an additional cost in terms of training time. Overall, the findings indicate that the KAN architecture is a promising approach for IMU-based inertial odometry applications in terms of compact model structure and reduced cumulative drift. Nevertheless, its generalization performance over different window lengths, sampling frequencies, and flight sequences should be investigated in more detail in future studies.